%% file: main.tex
\documentclass[letterpaper,twocolumn,10pt]{article}
\usepackage{usenix}
\usepackage{hyperref}

\usepackage{verbatim}
\usepackage{xspace}
\usepackage{tikz}
\usepackage{listings}
\usepackage{graphicx}
\usepackage{pgfplots}
\usepackage{tabularx, booktabs}
\usepackage{url}
\usepackage{algorithm}
\usepackage{algpseudocode}
\usepackage{amsmath}
\usepackage{multirow}
\usepackage{makecell}
\usepackage{enumitem}
\usepackage[export]{adjustbox}
\usepackage{subcaption} 

\newcommand{\spara}[1]{\vspace*{0.05in}\noindent\textbf{#1}\hspace{1mm}}
\newcommand{\addition}[1]{{#1}} 
\lstdefinelanguage{NextDoorAPI}
    {keywords={long, int, char, for, if, else, then, Vertex, Edge, Sampler, bool, return, while, true, float, true, false, void},
    keywordstyle=\textbf,
    sensitive=true,
    morecomment=[l][\color{greencomments}]{///},
    morecomment=[l][\color{greencomments}]{//},
    morecomment=[s][\color{greencomments}]{{(*}{*)}},
    morestring=[b]",
    stringstyle=\color{redstrings},
    escapeinside={(*}{*)}, 
    escapechar=|, xleftmargin=1.0ex,numbersep=4pt,
    emph={srcInit, steps, next, distinct, sampleSize, unique, stepTransits, samplingType},
    emphstyle={\color{red}},
    mathescape=true, 
    basicstyle=\ttfamily
    }

\begin{document}
\date{}
\title{Scalable Graph Neural Network Training: The Case for Sampling }

\author{ {\rm Marco Serafini, Hui Guan}\\
University of Massachusetts Amherst\\
United States
}


\maketitle

\begin{abstract}
    Graph Neural Networks (GNNs) are a new and increasingly popular family of deep neural network architectures to perform learning on graphs.
    Training them efficiently is challenging due to the irregular nature of graph data.
    The problem becomes even more challenging when scaling to large graphs that exceed the capacity of single devices.
    Standard approaches to distributed DNN training, such as data and model parallelism, do not directly apply to GNNs.
    Instead, two different approaches have emerged in the literature: \emph{whole-graph} and \emph{sample-based} training.
    
    In this paper, we review and compare the two approaches.
    Scalability is challenging with both approaches, but we make a case that research should focus on sample-based training since it is a more promising approach.
    Finally, we review recent systems supporting sample-based training.
\end{abstract}

\section{Introduction}

Many datasets are relational in nature: they are best represented as entities connected by relationships rather than as a single uniform dataset or table.
Graphs are a universal formalism to model relational data.
They are also commonly used to integrate data coming from multiple and possibly diverse data sources, such as relational databases, social networks, financial transaction logs, and many others.
Graph analysis can leverage relations to get a deeper insight into data.


Machine learning and deep learning are being increasingly used for  graph analytics.
Graph Neural Networks (GNNs), in particular, represent a new family of Deep Neural Network (DNN) architectures tailored for graphs, where the structure of the neural network overlaps with the structure of the graph itself~\cite{gnn-survey-comprehensive,gnn-survey-review}.
A GNN is composed of several layers and each layer transforms input features to output features. The output features from a GNN are usually referred to as \textit{embeddings} and used for downstream tasks.   
They are the state-of-the-art approach for several prediction and classification tasks on graphs.

Training GNNs presents unique challenges due to the irregularity of graph data.  
The input to a GNN is not a set of independent data items but a graph consisting of interconnected and inter-dependent vertices.
Each layer in GNN is thus modeled as a message-passing process whereby each vertex aggregates the features of its neighbors~\cite{dgl,pytorch-geometric}.
This results in an irregular and sparse computation, which is hard to perform efficiently. The problem becomes even more challenging when one wants to scale training to large graphs and multiple devices. 
Partitioning the graph inevitably splits some neighboring vertices across different partitions, leading to significant communication overhead.


Scaling GNN training is an open research question.
Frameworks designed for GNN training such as DGL~\cite{dgl} and PyG~\cite{pytorch-geometric} translate message-passing specifications of GNNs into those of DNN models, which are then run by existing DNN frameworks (e.g., TensorFlow~\cite{tensorflow} or PyTorch~\cite{pytorch}).
These DNN frameworks, however, are not specifically designed to scale GNNs to large input graphs.
Recent work has proposed dedicated techniques to scale GNN training that fall into one of two approaches: \emph{whole-graph} and \emph{sample-based} training.
In whole-graph training, message passing among vertices is performed on the entire graph.
Sample-based training first samples the graph to obtain mini-batches, then maps each mini-batch to one device, and finally performs training on each mini-batch independently.

In this paper, we describe and compare the two approaches from a system scalability perspective.
We discuss how scaling is challenging under both models, but  argue that sample-based training is a more promising approach.
Whole-graph training introduces inherent coordination and communication overheads that are hard to overcome as the system scales.
Scaling sample-based training, on the other hand, requires (a) sampling algorithms that can form mini-batches without incurring into the ``neighbor explosion'' problem~\cite{graphsage, samplingsurvey}
and (b) scalable systems to execute these sampling algorithms efficiently.
We review recent research that addresses these requirements.
Based on this review, we argue that sample-based training is a more promising approach to scaling GNN training.

\section{Background}

\spara{Graph Neural Networks}
GNNs perform \emph{representation learning}: they take a graph as an input and map each vertex to a $d$-dimensional vector, known as an \emph{embedding}.
Embeddings are then used as inputs for downstream machine learning tasks, such as vertex classification and link prediction.

GNN frameworks allow expressing models using the message-passing paradigm~\cite{dgl,pytorch-geometric}.
Each GNN layer can be expressed as a vertex-centric message-passing round.
At the $k$-th layer, each vertex $v$ aggregates the features $h^{(k-1)}_u$ of its neighbors $u \in N(v)$ and uses that information to compute its new feature $h^{(k)}_v$.
This happen through three functions: a message function $\phi$, a reduce function $\rho$, and an update function $\psi$.

Let $G = (V, E)$ be the input graph.
The message function computes, for each edge $e$, a message from the source to the destination vertex given the features of the incident vertices and the edge:
$$
    m^{(k)}_e = \phi(h^{(k-1)}_u, h^{(k-1)}_v) \quad \forall e=(u,v) \in E
$$

The message function can simply output the features of the source vertex $h^{(k-1)}_u$.
Some algorithms assign features also to edges, and use those features as input to the message function.

The reduce function aggregates multiple messages sent to the same receiving vertex, for example by summing them or using an LSTM network:
$$
    a^{(k)}_v = \rho(\{m^{(k)}_e: e=(u,v) \in E\}) \quad \forall v \in V
$$

Finally, the update function takes the result of message aggregation and computes a new feature vector by applying DNN operators such as fully-connected layers or convolutions:
$$
    h^{(k)}_v = \psi(h^{(k-1)}_v, a^{(k)}_v) \quad \forall v \in V
$$

By using $n$ GNN layers, the output features of each vertex (also called embedding) can reflect features from all its $n$-hop neighbors.
The three functions $\phi$, $\rho$, and $\psi$ constitute the GNN model and encapsulate its parameters.


\spara{Distributed DNN Training}
Before discussing approaches to scale GNN training, it helps to review the basic approaches to scale  DNN training that are adopted by existing frameworks such as TensorFlow~\cite{tensorflow} and PyTorch~\cite{pytorch}.
Scalability requires distributing the workload across devices, typically GPUs, that execute the DNN model.
Data and model parallelism are the most common paradigms to do large-scale DNN training.

\emph{Data parallelism} partitions the input data into mini-batches and assigns each mini-batch to one device. 
Each device has a full copy of the model and it independently performs iterations over the model (forward and backward propagation) using the mini-batch as input.
The iteration produces gradients for the model parameters.
Gradients from all devices are then aggregated, using a parameter server~\cite{parameter-server} or an all-reduce protocol~\cite{allreduce}, and finally applied to each copy of the model.

Data parallelism is the default choice to parallelize training because each device can complete iterations independent of each other.
It is particularly suitable for small models, such as models with small convolutional filters, because the cost of aggregating gradients across devices is limited~\cite{owt}.
However, for large models, gradient aggregation can be a substantial overhead.
Furthermore, if the model is too large or complex, it might exceed the capacity of one device, making data parallelism insufficient.

\emph{Model parallelism} partitions the model across multiple devices, for example by assigning a subset of consecutive layers to the same device.
Completing an iteration requires interaction among devices, which exchange the features produced by their partition of the model.
In this case, the model parameters are not replicated and gradients can directly be applied by the device that computes them.

Model parallelism is the best fit for models that exceed the capacity of one device.
They also perform well when each partition of the model requires a large amount of computation and produces small features.
It does not require exchanging gradients, so it works well with models with a large number of parameters, such as models with many fully-connected layers~\cite{owt}.
An important drawback of model parallelism is that it requires tighter coordination among devices: downstream devices can only make progress once upstream devices send their features.
The communication and coordination cost of model parallelism can be reduced by using techniques like pipelining that combine mini-batching and model partitioning~\cite{pipedream}.


\section{Scaling GNN Training}

\begin{figure}[t]
\centering
\includegraphics[width=0.5\textwidth]{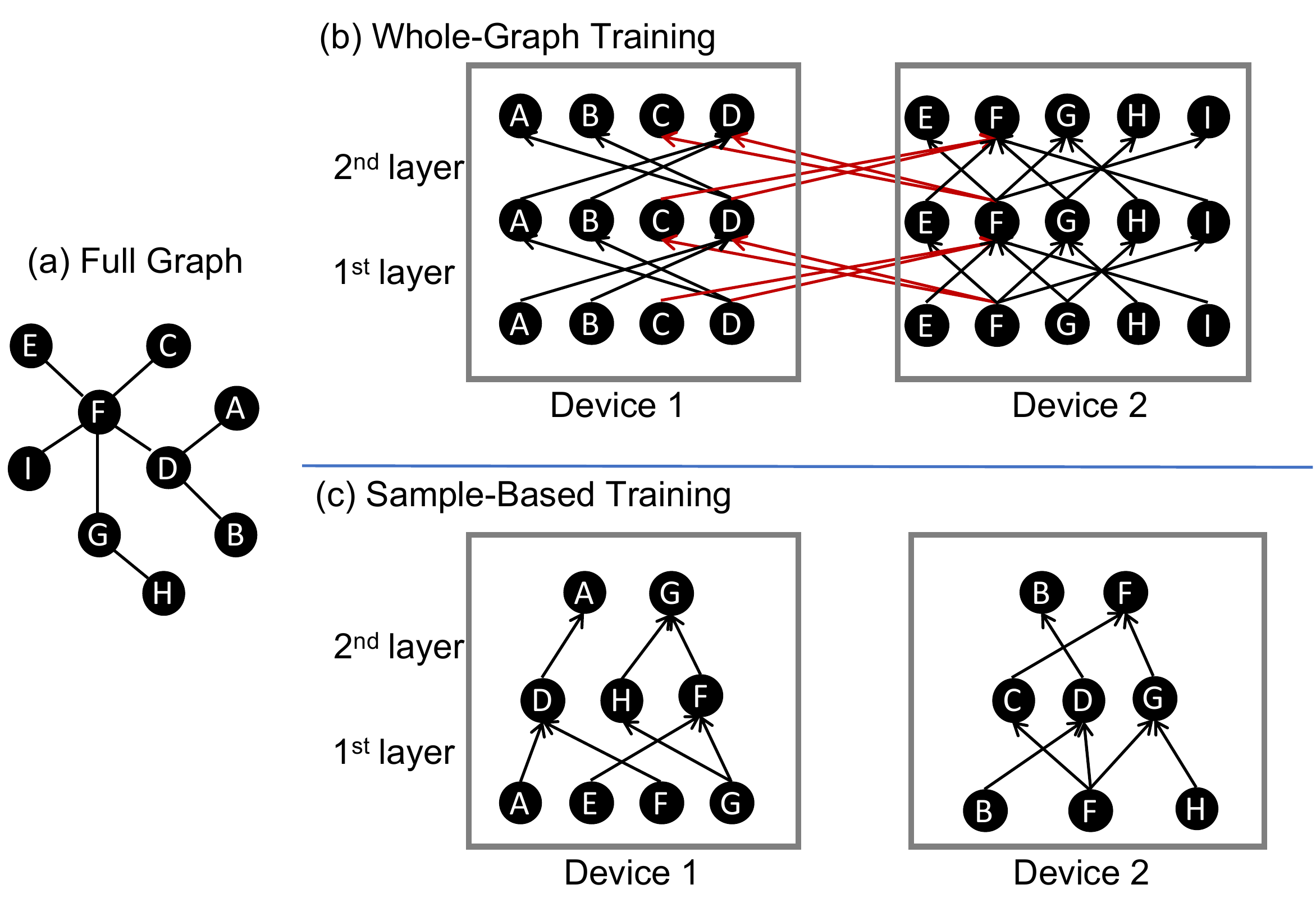}
\caption{Example of whole-graph and sample-based training. }
\label{fig:whole-vs-sample}
\end{figure}

When the input graph is too large to be handled by one device, it is necessary to use a distributed approach.
However, neither data nor model parallelism are a good fit for GNN training.
A graph is a single data structure consisting of interconnected vertices, not a collection of small independent samples, as typically assumed in data parallelism.
There is no way to partition the vertices of most graphs without having edges between partitions.
As discussed, on the other hand, GNN models are relatively shallow and small compared to DNNs, making model parallelism a bad fit for GNN as well.


Because of the uniqueness of GNNs, prior work has proposed two approaches that are tailored to distributed GNN training: \emph{whole-graph} and \emph{sample-based} training.
We now present and compare them.
Table~\ref{tab:comparison} reports a summary of the comparison and Figure~\ref{fig:whole-vs-sample} shows an example of a two-layer GNN trained on two devices using the two approaches.

\input{tables/comparison.tex}

\spara{Whole-Graph Training}
Whole-graph training was adopted by two systems for GNN training, NeuGraph~\cite{ma2019neugraph} and Roc~\cite{roc}.
It partitions the input graph and assigns each partition to one device.
One way to look at whole-graph training is to treat the input graph as a single sample, which is then partitioned across one of the attribute dimensions~\cite{roc}. 
The GNN model parameters are replicated on all devices.
Each iteration is processed one GNN layer at a time.
At each layer $k$, each device computes the feature vector $h^{(k)}_v$ for each vertex $v$ in its partition.
This requires fetching the feature $h^{(k-1)}_u$ for each neighbor $u \in N(v)$ of $v$, which typically include features computed by other devices in the previous layer.
Therefore, in whole-graph training devices must exchange features to complete iterations, a characteristic whole-graph training shares with model parallelism.
Whole-graph training is a form of full-batch training: each iteration is executed on the entire input.

In the example of Figure~\ref{fig:whole-vs-sample}, whole-graph training partitions the vertices on the left side across two devices.
The edges marked in red connect vertices in two different partitions.
At every layer, each vertex must aggregate the features of its neighbors, which creates cross-device communication and coordination along those edges.
In the forward pass, devices exchange features (i.e., the vertex features) at each layer, whereas in the backward pass they exchange the gradients of the features.

\spara{Sample-Based Training}
Sample-based training is supported by AliGraph~\cite{aligraph}, DistDGL~\cite{distdgl}, and PaGraph~\cite{pagraph}.
The first two target distributed training, the last one runs on multi-GPU systems.
This approach first performs graph sampling to create mini-batches (also referred to as \textit{samples}), and then trains on the mini-batches in parallel on different devices.
The goal of sample-based training is to enable devices to complete iterations independently on their mini-batches without having to exchange features.
Each mini-batch includes the data necessary to compute the output feature vector for some vertices called \textit{target vertices}. 
In a GNN model with $n$ layers, each mini-batch includes the input features of the $n$-hop neighborhood of those target vertices.
Sample-based training is a form of mini-batch training since iterations are computed on a piece of graph data.

This approach differs from data parallelism since mini-batches contain redundant data.
Each vertex is included in the $n$-hop neighborhoods of multiple vertices, and can hence be included in multiple mini-batches.
Nonetheless, this approach has analogies to data parallelism.
Devices can complete iterations without communicating with each other.
Furthermore, exchanging the gradients among devices is not expensive since the model is small.

Consider again the example of Figure~\ref{fig:whole-vs-sample}.
Sample-based training creates two mini-batches and assigns them to the two devices.
The first device computes the output features of target vertices A and G, whereas the second one computes them for B and F.
Each mini-batch includes all vertices in the two-hop neighborhood of the target vertices assigned to the device.
Sampling the graph to obtain mini-batches is part of the training process but it is not depicted in the figure.

\spara{Why Scaling Whole-Graph Training is Difficult}
The main drawback of whole-graph training is that devices need to exchange features (the updated vertex features) at each GNN layer (see Figure~\ref{fig:whole-vs-sample}(b)).
The communication complexity depends on the number of edges across partitions and can be large.
Exchanging features also introduces control dependencies between devices within the context of one iteration.

Data-dependent communication and coordination constraints among devices are a major hurdle when scaling to large clusters.
Increasing the number of partitions reduces the computational load per worker but it also increases the number of edges across partitions, resulting in higher communication and coordination costs both globally and locally at each device.

These scalability bottlenecks are not unique to whole-graph GNN training.
Distributed graph computation systems such as Pregel also partition vertices among workers~\cite{pregel,powergraph,PowerLyra,gemini}.
Each worker must coordinate with the others at each superstep and exchange messages among neighboring vertices.
There are important differences between the two classes of systems. 
In whole-graph GNN training, devices exchange known data types (tensors) and perform known operators, for example sparse matrix multiplication.
This enables more fine-grained vectorization and scheduling compared to graph processing systems, where vertices execute UDFs and exchange messages that are opaque to the system.
However, the communication and coordination patterns are similar.
Scaling out graph computation systems to large clusters has always been a difficult research challenge (see for example~\cite{ldbc-graphalytics}) and the same holds for whole-graph GNN training.

Moving towards sample-based training for better scalability does not necessarily means degraded model accuracy. Although some recent system work on improving GNN training scalability point out that sampling techniques come with potential model accuracy loss for large real-world graphs, they compare whole-graph training with few graph sampling approaches~\cite{roc, ma2019neugraph}. The more recent graph sampling literature has consistently shown better accuracy on common graph benchmarks~\cite{clustergcn, graphsaint}. 


\spara{A Case for Sample-Based Training}
In this paper, we claim that sample-based training is a more promising approach to distributed GNN training.
Instead of revisiting known scalability bottlenecks in the new context of GNNs, we should leverage the new opportunities offered by GNNs to avoid those bottlenecks altogether.

Sample-based training eliminates per-layer coordination and communication costs.
Devices only need to exchange gradients, which is a small cost since the size of the model is small and independent of the size of the graph.
Using asynchrony or bounded-staleness for gradient exchange can further reduce these coordination costs.

Sample-based training also results in a more modular system design.
Once devices obtain mini-batches, they can calculate gradients by using any GNN training frameworks designed for graphs that fit in one device.
Gradient exchange can occur just like in data parallelism, which is already well supported.

This does not mean that scaling sample-based training is straightforward.
There are two main challenges to scalability: redundant work and the sampling overhead.
However, recent research indicates that both challenges can be solved.

Redundant work arises from the ``neighbor explosion'' problem: each vertex can have a very large number of $n$-hop neighbors, which must be included in its mini-batch.
Multiple mini-batches are likely to overlap in many common vertices.
Multiple devices must then compute the features for these vertices, resulting in redundant computation and memory costs.
For example, in Figure~\ref{fig:whole-vs-sample}, the first-layer feature of vertex D, $h^{(1)}_D$, is computed by both devices, and the input features of vertex A are replicated at both devices.

A solution to this problem is to sub-sample the subgraphs constituting the mini-batches to prune some vertices and edges.
Re-sampling mini-batches at each iteration reduces the chances of missing important information.
Developing sampling algorithms to create mini-batches that maximize accuracy and minimize training time is a very active area of research and has shown that this is feasible.
Table~\ref{tab:end-to-end} lists some of these algorithms, which we discuss further in Section~\ref{sec:algos}.

Graph sampling can take a significant portion of the total training time in real-world applications.
The computation is irregular and is typically performed using the CPU.
In our previous work, we found that graph sampling can take up to 62\% of an epoch's time if the host has a single GPU (see Table~\ref{tab:sampling-overhead})~\cite{nextdoor}.
This bottleneck is further exacerbated if the CPU is attached to multiple GPUs consuming samples for training.
Therefore, speeding up and scaling sampling is an important problem for graph sampling.
Preliminary work on this front shows promising results too, as we will discuss in Section~\ref{sec:systems}.

\input{tables/sampling-cost.tex}

\section{Scaling Sample-Based Training}
In this section, we review recent work on graph sampling for GNN training.
We first describe algorithms to obtain the subgraphs that constitute the mini-batches of sample-based training.
Then, we discuss systems to speed up the execution of sampling algorithms.

\subsection{Sampling Algorithms}
\label{sec:algos}
Sampling algorithms in GNN training aim to select a subset of vertices and edges based on certain rules. After sampling, instead of using all neighbors as in the whole-graph training, sample-based training constructs a vertex's feature by only aggregating the features of the sampled set of neighbors. 
Existing sampling approaches largely fall into four categories: \textit{node-wise} sampling, \textit{layer-wise} sampling, \textit{subgraph-based} sampling, and \textit{heterogeneous} sampling~\cite{samplingsurvey}. They differ in the granularity of the sampling operation in one training mini-batch. \textit{Heterogeneous} sampling applies to heterogeneous graphs whose edges and vertices are of different types. As our discussion focuses on homogeneous graphs, we next elaborate on the first three types of sampling algorithms.

\textit{Node-wise} sampling applies sampling operations to each vertex's neighbors: a part of neighbors of a vertex are sampled based on specific probability (e.g., uniform distribution) to compute the vertex's feature. One typical example is GraphSAGE~\cite{graphsage}. It uniformly  samples a fixed number of neighbors for each vertex in the graph and aggregates their features to generate the vertex's feature in each GNN layer. The output feature of each vertex from the final GNN layer is then used for the GNN model's weight update and downstream tasks. Its variants include PinSage~\cite{pinsage}, SSE~\cite{sse}, VR-GCN~\cite{vrgcn}, and  MVS~\cite{mvs}, which differ in the design of sampling functions. 

\textit{Layer-wise} sampling samples multiple vertices simultaneously for each GNN layer in one sampling step. This approach is usually faster than node-wise sampling as it avoids the exponential extension of neighbors. Example algorithms include FastGCN~\cite{fastgcn} which samples a fixed number of vertices in each GNN layer based on pre-calculated probability independently and LADIES~\cite{ladies} and AS-GCN~\cite{asgcn} which introduce layer dependencies and sample vertices in the $k$-th GNN layer based on vertices sampled in the $k+1$-th layer.

\textit{Subgraph-based} sampling samples a subgraph, which is composed of selected vertices and edges, and conducts training using the subgraph.  Existing work generate subgraphs by either partitioning the whole graph or extending vertices and edges using specific policies~\cite{clustergcn, graphsaint, bai2020ripple}. For example, ClusterGCN~\cite{clustergcn} first partitions the whole graph into multiple clusters using graph clustering algorithms and then randomly samples a fixed number of clusters as a mini-batch by combining these clusters into a subgraph.  
GraphSAINT~\cite{graphsaint}, on the other hand, leverages random walk to sample neighbors of a vertex and generate subgraphs with the selected vertices.  


\subsection{Systems for Efficient Sampling}
\label{sec:systems}
Building samples of a graph is an irregular computation that is hard to perform efficiently.
Graph sampling algorithms are designed to preserve good accuracy while generating small mini-batches.
The researchers who develop these algorithms should not have to deal with the low-level details of hardware architectures to optimize performance.
There is an emerging need for systems that can abstract away the graph sampling process by offering a high-level yet generic API and an efficient runtime to execute these programs.
These systems should be able to scale to large graphs and integrate with the training process.

\input{tables/nd-e2e-speedups.tex}

\spara{Fast Graph Sampling on GPUs}
Sample-based GNN implementations often rely on frameworks such as DGL and PyTorch Geometric to perform GNN training on GPUs.
However, they perform graph sampling on the CPU.
This is in part due to the lack of frameworks for efficient and generic sampling on GPUs.

Our recent work on NextDoor addresses this problem~\cite{nextdoor}.
NextDoor enables users to express graph sampling tasks using a general, high-level API.
It then executes these tasks efficiently using GPUs.
The API is expressive enough to support the sampling algorithms described previously.
Using NextDoor can speed up the end-to-end training time of existing GNN systems by up to 4.75$\times$ (see Table~\ref{tab:end-to-end}).

\input{figures/pseudocode.tex}

Figure~\ref{fig:2-hop-sage} shows an implementation of a sampling algorithm using the NextDoor API.
The algorithm is used by GraphSAGE~\cite{graphsage}, a classic GNN algorithm.
This particular implementation samples the 2-hop neighborhood of a vertex.
The \texttt{steps} function indicates that sampling performs two recursive hops from the vertex.
The \texttt{sampleSize} function returns how many neighbors of a vertex are sampled at each step: 25 neighbors of the starting vertex in the first step and 10 neighbors of each sampled neighbor in the second. 
The \texttt{next} function specifies how to pick a neighbor of a vertex.

Beyond offering an easy-to-use API, NextDoor introduced a novel approach to parallelize graph sampling called \emph{transit parallelism}.
This is better suited to GPU architectures than the approach used by other systems for graph sampling, such as KnightKing~\cite{knightking} and C-saw~\cite{c-saw}, and graph mining, such as Arabesque and others~\cite{arabesque, asap, rstream, fractal, gminer, peregrine, pangolin, automine}.
All these systems expand multiple samples in parallel and assign each sample to a group of consecutive threads, which could be part of the same warp.
This approach, which we call \emph{sample parallelism}, results in sub-optimal performance on GPUs because of their hardware architecture.
GPU kernels achieve optimal performance when threads in a warp access contiguous memory locations, cache data on shared memory, and perform the same steps. 
Sample parallelism does not have these properties.

To see why, consider the example of Figure~\ref{fig:whole-vs-sample}, and suppose that the same GPU is computing the two mini-batches for the two devices.
Each mini-batch corresponds to a sample.
Suppose that the algorithm is now expanding the first sample by adding neighbors of its vertices A and G.
A sample-parallel execution of the sampling algorithm would associate the sample to a group of consecutive threads, which are likely to be in the same warp.
These threads will access the adjacency lists of two different vertices, A and G.
This leads to uncoalesced memory accesses and poor access locality since the two adjacency lists are located at random locations in the GPU memory.
It also results in warp divergence if the threads scan adjacency lists of different sizes.

NextDoor uses each \emph{transit vertex} as the fundamental unit of parallelism when building the samples.
A transit vertex is a vertex whose neighbors may be added to one or more samples of the graph. 
NextDoor groups all samples that need to ``transit across'' a vertex and assigns the samples to consecutive GPU threads.
Each thread accesses the adjacency list of the transit and adds one neighbor of the transit vertex to one sample.
Since all threads access the same list, NextDoor achieves  coalesced global memory reads, low warp divergence, and effective caching using the shared memory of the GPU.
We compared two versions of NextDoor that parallelize sampling by sample and by transit, using several sampling algorithms implemented using NextDoor's API.
Transit parallelism has shown to be consistently faster, as shown in~\cite{nextdoor}.

Consider again the example of Figure~\ref{fig:whole-vs-sample} and suppose that the same GPU is computing the mini-batches (i.e., the samples) for the two devices.
Suppose that the two samples already contain the target vertices (\{A, G\} and \{B, F\} respectively) and their one-hop neighbors (\{D, H, F\} and \{C, D, G\} respectively).
The algorithm is now expanding both samples by adding the two-hop neighbors of the target vertices.
Vertex D is a transit vertex for both samples because the algorithm must select neighbors of D for both samples.
NextDoor assigns a group of consecutive threads to vertex D and both samples. 
All these threads read the adjacency list of D.
Accesses to the list can be coalesced and the list can be cached in shared memory.
If the threads scan the adjacency list, they will perform the same number of steps.
These properties result in better performance on GPUs because of their hardware architecture.


\spara{Scaling to Larger Graphs}
Scaling sample-based training to large graphs critically relies on scaling sampling.
Once the sampling process generates mini-batches, distributed training on each mini-batch proceeds independently similar to data parallelism.
However, efficiently sampling large graphs that are stored on a distributed system is non-trivial.

One research question to address is how to scale graph  sampling across devices. 
Some existing systems have already explored the problem. 
KnightKing computes random walks on distributed graphs and uses a graph processing system as a substratum~\cite{knightking}.
AliGraph~\cite{aligraph} and DistDGL~\cite{distdgl} are end-to-end GNN training systems, integrating sampling and training.
PaGraph performs sample-based GNN training in multi-GPU systems~\cite{pagraph}.
KnightKing and DistDGL run several sampling workers that incrementally expand their samples and pull remote graph data on demand.
Distributed graph mining systems like G-Miner use a similar approach~\cite{gminer}.
These systems, however, differ in the way they schedule distributed sampling.
Finding the optimal strategy is still an area for research, and it will be interesting to see whether ideas from graph mining systems can be borrowed in this context.

Another problem with the aforementioned systems is that they perform sampling using CPUs only.  
GPU-based sampling has the potential to significantly reduce end-to-end training time, as shown in NextDoor~\cite{nextdoor}.
Distributed sampling systems can leverage GPUs for improved throughput. 
NextDoor supports multi-GPU--based sampling if the graph fits in the device memory.
How to scale the process to larger graphs is still an open area of research. 
Given that GPUs are also used for GNN training, another open issue is how to efficiently overlap sampling with training on the same set of devices. 

Finally, how to distribute and store the graph data and how to transfer it in and out of the GPUs is another critical aspect in the performance of distributed sample-based training.
PaGraph, for example, uses caching to minimize data transfers\cite{pagraph}.
More research is likely to unveil additional optimizations.




\section{Conclusions}
In this paper, we have compared two approaches to scale GNN training: whole-graph and sample-based.
Whole-graph training requires devices to coordinate and communicate at each GNN layer, which represents a challenge for scalability.
Sample-based training avoids this coordination altogether, so it promises better scalability.
Scaling sample-based training requires (a) sampling algorithms that can form mini-batches without losing too much information or generating excessive redundant work, and (b) systems that can execute these algorithms efficiently.
Recent work indicates that both requirements can be fulfilled.

\section*{Acknowledgements}
The authors would like to thank the NextDoor team: Abhinav Jangda, Sandeep Polisetty, and Arjun Guha.
This work was partially supported by a Facebook Systems for Machine Learning Award and an AWS Cloud Credit for Research grant.

\bibliographystyle{plain}
\bibliography{main}

\end{document}

%% file: tables/comparison.tex
\begin{table*}
    \small
    \centering
    \begin{tabularx}{\linewidth}{c||c|c|c|c|c|c}
        \toprule
        \textbf{Model}      & \textbf{Unit of}      & \textbf{Gradient}    & \textbf{Exchange of} & \textbf{Exchange of}   & \textbf{Redundant}  & \textbf{External overheads} \\
                            & \textbf{parallelism}  & \textbf{descent}     & \textbf{activations} & \textbf{gradients}     & \textbf{computation} & \\\hline
        Whole-graph         & Vertex-centric        & Full-batch           & Yes (per-layer)      & Yes                    & No                 & Graph partitioning (per-task)\\
        Sample-based      & Subgraph-centric      & Mini-batch           & No                   & Yes                    & Yes               & Sampling (per-iteration)\\
    \bottomrule
    \end{tabularx}
    \caption{Comparison of approaches to scalable GNN training.}
    \label{tab:comparison}
\end{table*}

%% file: tables/sampling-cost.tex
\begin{table}
    \small
    \centering
  
    \begin{tabularx}{0.57\linewidth}{l|c|c}
      \toprule
     Input Graphs                 &  PPI     & Reddit \\\hline
    GraphSAGE~\cite{graphsage}    &  51\%   & 45\%   \\
    FastGCN~\cite{fastgcn}        &  26\%    & 52\%\\
    LADIES~\cite{ladies}          &  40\%    & 62\%\\
    ClusterGCN~\cite{clustergcn}  & 4.1\%    & 24\%\\
    GraphSAINT~\cite{graphsaint}  & 25\%     & 30\% \\
    MVS~\cite{mvs}                & 24\%     & 25\%\\
    \bottomrule
    \end{tabularx}
    \caption{Fraction of time spent in graph sampling and training by different GNN algorithms (from~\cite{nextdoor}).}
    \label{tab:sampling-overhead}
\end{table}

%% file: tables/nd-e2e-speedups.tex
\begin{table}[t]
    \small
    \centering
    \begin{tabularx}{\linewidth}{c|ccccc}
      \toprule
              & PPI           & Reddit       & Orkut & Patents      & LiveJ \\
    \hline
    GraphSAGE & 1.30$\times$  & 1.21$\times$ & OOM   & 1.20$\times$ & 1.22$\times$\\
    \addition{FastGCN}   & \addition{1.25$\times$}  & \addition{1.52$\times$} & \addition{4.75$\times$}  & \addition{2.3$\times$}  & \addition{4.31$\times$}\\
    \addition{LADIES}    & \addition{1.07$\times$}  & \addition{1.37$\times$} & \addition{2.27$\times$}  & \addition{2.1$\times$}  & \addition{2.34$\times$}\\
    \addition{ClusterGCN} &\addition{ 1.03$\times$} & \addition{1.20$\times$} & \addition{OOM} & \addition{1.4$\times$} & \addition{1.51$\times$}\\
    \bottomrule
    \end{tabularx}
    \caption{\addition{End-to-end speedups after integrating NextDoor in GNNs over vanilla GNNs  (from~\cite{nextdoor}). \label{tab:end-to-end}}}
\end{table}

%% file: figures/pseudocode.tex
\begin{figure}
    \footnotesize
    \begin{lstlisting}[language=NextDoorAPI]
Vertex next(s, transits, srcEdges, step) {|\label{line:sage:begin}|
    int idx = randInt(0, srcEdges.size());
    return srcEdges[idx];} |\label{line:sage:end}|
int steps() {return 2;} |\label{line:sage:steps}|
int sampleSize(int step) {
    return (step == 0) ? 25 : 10;}|\label{line:sage:maxsize}|
bool unique(int step) {return false;}
SamplingType samplingType(){
    return SamplingType::Individual;}
Vertex stepTransits(step, s, transitIdx){
    return s.prevVertex(1, transitIdx);}
            \end{lstlisting}
        \caption{GrapSAGE's 2-hop neighbors sampling implemented using the NextDoor API (from~\cite{nextdoor}).
        }
        \label{fig:2-hop-sage}
\end{figure}